%% file: main.tex
\documentclass[a4paper]{article}   

\input{math_commands.tex}

\usepackage{hyperref}
\usepackage{url}
\usepackage{booktabs}       
\usepackage{amsfonts}       
\usepackage{nicefrac}       
\usepackage{microtype}      

\input{custom}

\title{\LARGE \bf
Efficient Reinforcement Learning in Resource Allocation Problems
  Through Permutation Invariant Multi-task Learning
}

 \author{Desmond Cai,$^{1}$ Shiau Hong Lim$^{2}$,  Laura Wynter$^{2}$
 \thanks{$^{1}$Desmond Cai is with AStar, Singapore
        {\tt\small desmond.cai@gmail.com}}%
   \thanks{$^{2}$Shiau Hong Lim and Laura Wynter are with IBM Research, Singapore
        {\tt\small \{shonglim,lwynter\}@sg.ibm.com}}%
  }

\begin{document}

\maketitle

\begin{abstract}
    One of the main challenges in real-world reinforcement learning is to learn successfully from limited training samples. We show that in certain settings, the available data can be dramatically increased through a form of multi-task learning, by exploiting an invariance property in the tasks. We provide a theoretical performance bound for the gain in sample efficiency under this setting. This motivates a new approach to multi-task learning, which involves the design of an appropriate neural network architecture and a prioritized task-sampling strategy. We demonstrate empirically the effectiveness of the proposed approach on two real-world sequential resource allocation tasks where this invariance property occurs: financial portfolio optimization and meta federated learning.
\end{abstract}

\input{introduction}
\input{algorithm}

\input{experiment}

\section{Conclusions}
We  introduce an approach for increasing the sample efficiency of reinforcement learning in a  setting with  widespread applicability within the class of sequential resource allocation problems. This property is permutation invariance:  resources are allocated to   entities according to a score, and the order can change without modifying the optimal allocation. Under this property, we show that a bound exists on the policy performance. This bound motivates a highly effective algorithm for improving the policy through a multi-task approach. Using  prioritized task-sampling, the method  not only  improves the reward of the final policy but also  renders it more robust. We illustrate the property and the method on two important problems: sequential financial portfolio optimization and meta federated learning, where the latter is provided in the Appendix. 


\section*{Appendix}
\input{supp}
\bibliographystyle{plain}
\bibliography{references}

\end{document}

%% file: math_commands.tex
\usepackage{amsmath,amsfonts,bm}

\def\eqref#1{equation~\ref{#1}}

\def\1{\bm{1}}

\DeclareMathAlphabet{\mathsfit}{\encodingdefault}{\sfdefault}{m}{sl}
\SetMathAlphabet{\mathsfit}{bold}{\encodingdefault}{\sfdefault}{bx}{n}



%% file: custom.tex
\usepackage{amsmath}
\usepackage{amssymb}
\usepackage{mathtools}
\usepackage{algorithm}
\usepackage{algorithmic}

\newenvironment{proof}{\paragraph{Proof:}}{\hfill$\square$}

\newtheorem{theorem}{Theorem}
\newtheorem{cor}{Corollary}
\newtheorem{assumption}{Assumption}
\newtheorem{definition}{Definition}
\newtheorem{remark}{Remark}

\usepackage{color}
\usepackage{ifthen}
\definecolor{darkgreen}{rgb}{0,0.5,0}
\newboolean{showcomments}
\setboolean{showcomments}{true}
\newcommand{\desmond}[1]{\ifthenelse{\boolean{showcomments}}{\textcolor{red}{(Desmond says: #1)}}{}}
\newcommand{\shiau}[1]{\ifthenelse{\boolean{showcomments}}{\textcolor{red}{(Shiau Hong says: #1)}}{}}
\newcommand{\laura}[1]{\ifthenelse{\boolean{showcomments}}{\textcolor{red}{(Laura says: #1)}}{}}

%% file: introduction.tex
\section{Introduction}

Sample efficiency in reinforcement learning (RL) is an elusive goal. Recent attempts at increasing the sample efficiency of RL implementations have focused to a large extent on incorporating models into the training process: \cite{model1, model2, model3, model4, model5, model6, model7, model8, model9}. The models encapsulate knowledge explicitly, complementing the experiences that are gained by sampling from the RL environment. Another means towards increasing the availability of samples for a reinforcement learner is by tilting the training towards one that will better transfer to related tasks: if the training process is sufficiently well adapted to more than one task, then the training of a particular task should be able to benefit from samples from the other related tasks. This idea was explored a decade ago in \cite{lazaric} and has been gaining traction ever since, as researchers try to increase the reach of deep reinforcement learning from its comfortable footing in solving games outrageously well to solving other important problems.  \cite{towards} discusses a number of methods for increasing sample efficiency in RL and includes experience transfer as one  important avenue, covering the transfer of samples, as we do here,  transfer of representation or skills, and jumpstarting models which are then ready to be quickly, i.e. with few samples, updated to different tasks. 
\cite{deramo2020learning} address the same idea, noting that multi-task learning can  improve the learning  of each individual task, motivated by robotics-type tasks  with underlying commonality, such as balancing a single vs. a double pendulum, or hopping vs. walking.  

We are  interested  in exploiting the ability of multi-task learning  to solve the sample efficiency problem of RL. Our setting does not apply  to all problem classes nor does it seek to exploit the kind of physical similarities found in robotics tasks that form the motivation of \cite{lazaric, deramo2020learning}.
Rather, we show that there are a number of reinforcement learning tasks with a particular fundamental property that makes them ideal candidates for multi-task learning with the goal of increasing the availability of samples for their training. We refer to this property as permutation invariance. It is present in  very diverse tasks: we illustrate it on a financial portfolio optimization problem, whereby trades are executed sequentially over a given time horizon, and on the problem of meta-learning in a federated supervised learning setting. 

Permutation invariance in the financial portfolio problem exhibits itself as follows: consider the task of allocating a portion of wealth to each of a number of financial instruments using a trading policy. If the trading policy is permutation invariant, one can change the order of the instruments without changing the policy. This allows one to generate multiple portfolio optimization tasks from a given set of financial instruments.
A commonality between applications that have this property is that they concern sequential resource allocation: at each time step, the resource allocation scores the quality of each available candidate entity (for example a financial instrument in the above example),  then based on those scores, apportions out  the resource (the total wealth to invest, in the above example) among the entities at that time step, so that over the horizon of interest, the reward is maximized. 

Sequential resource allocation problems  include applications such as sequential allocation of budget, sequential allocation of  space, e.g. in IT systems, hotels, delivery vehicles,  sequential allocation of people to work slots or appointments, etc. 
Many such applications possess permutation invariance in that the ordering of the  entities, i.e. where the resources  are  allocated, can change without changing the resulting optimal allocation. 
We show that under this form of permutation invariance, it is possible to derive a bound on the performance of the policy. The bound is  an extension of that of \cite{lazaric}, and while similar to, provides additional information beyond  the bound of \cite{deramo2020learning}. We  use the bound to motivate an algorithm that allows for substantially improved results as compared with solving each task on its own. The bound and the algorithm are first analyzed on a synthetic problem that validates the bound  in our theorem and confirms the multi-task gain that the theory predicts.
\cite{popart, attentive} have cautioned against degrading of the performance on each task when some tasks bias the updates to the detriment of others in multi-task learning. They claim that some tasks have a greater density or magnitude of in-task rewards and hence  a disproportionate impact on the learning process. 
In our setting, deleterious effects of some tasks on others could also arise. The algorithm we propose handles this  through a form of prioritized sampling, where  priorities are put on the tasks themselves, and acts  like a prioritized experience replay buffer, applied to a multi-task learning problem. We show empirically that the priorities thus defined  protect the overall learning problem from the deleterious effects that unrelated or unhelpful tasks could otherwise have on the policy.

The contributions of this work are as follows: (1) we identify the permutation invariance property of the class of 
reinforcement learning problems involving sequential resource allocation, (2)  we define 
 a method to increase sample efficiency in these reinforcement learning problems by leveraging this property of permutation invariance; (3) we provide a theoretical performance bound for the class of problems; (4) we validate experimentally the utility of  permutation variance on sample efficiency as well as the validity of the bound on a synthetic problem;  and (5) we illustrate two real-world RL resource allocation tasks for which this property holds and demonstrate  the benefits of the proposed method on   sample efficiency and thus also on the overall performance of the models.

\section{Related work}

A notable first stream of work on leveraging multi-task learning for enhancing RL performance on single tasks can be found in \cite{ wilson, lazaric} which consider, as we do, that there is an underlying MDP from which the multiple tasks can be thought to derive. They use however a Bayesian approach and propose a different algorithmic method than ours. Our results extend performance bounds by
\cite{lazaric2012finite} on single-task RL.
As noted by \cite{towards}, jumpstarting, or distilling experiences and representations of relevant policies is another means to increasing sample efficiency in solving a new but related problem. \cite{rusu} uses this idea in so-called progressive neural networks and \cite{parisotto} leverage multiple experts to guide the derivation of a general policy. With a similar objective, \cite{teh} define a policy centroid, that is, a shared distilled policy, that captures the commonalities across the behaviors in the tasks. In all of these distillation-type methods, the tasks considered are simple or complex games. 

\cite{teh} note that their policy centroid method, distral, is likely to be affected by task interference, in that  differences across tasks  may degrade the performance of the resulting policy of any of the constituent tasks. This topic was studied by  \cite{popart, attentive}. \cite{popart}  proposed a solution to this by extending the so-called PopArt normalization \cite{popart1} to re-scale the updates of each task so that the different characteristics  of the task-specific reward do not skew the learning process. \cite{attentive} use a different approach that learns attention weights of the sub-networks of each task and discards those that are not relevant or helpful.
\cite{sharing_ijcai, deramo2020learning} are, like our work, concerned with sharing of experiences to facilitate a more sample-efficient learning process. \cite{sharing_ijcai} suggest  identifying the shared portions of tasks to allow sharing of samples in those portions. The work of \cite{deramo2020learning} is in some ways quite similar to ours: the authors' goal is the same and they derive a bound as we do on the performance in this setting. However, their setting is  different in that their tasks have both shared and task-specific components, and their bound  becomes tighter only as the number of tasks increases. In our setting, we do not require a task-specific component, and we are able to show how the distance between the MDPs of each task, in addition to the number of tasks, affects the strength of the bound.
Recently, permutation invariance has been exploited
in deep multi-agent reinforcement learning \cite{liu2019} where the invariance properties arise
naturally in a homogeneous multi-agent setting. Their work employs permutation invariance in learning the critic
whereas in our case the entire learned policy employs permutation invariance.

%% file: algorithm.tex
\section{Preliminaries}

We begin by
 defining  notation. 
For a measurable space with domain $\mathcal{X}$, 
let $\mathcal{S}(\mathcal{X})$ denote
the set of probability measures over $\mathcal{X}$, 
and $\mathcal{B}(\mathcal{X};L)$ 
the space of bounded measurable functions with domain $\mathcal{X}$ and bound $0 < L < \infty$.
For a measure $\rho\in\mathcal{S}(\mathcal{X})$ and a measurable function $f:\mathcal{X}\rightarrow\mathbb{R}$, 
the $l_2(\rho)$-norm of $f$ is $\|f\|_\rho$, and
for a set of $n$ points $X_1,\cdots,X_n\in\mathcal{X}$, 
 the empirical norm, $\|f\|_n$ is
\begin{align*}
    \|f\|_\rho^2 = \int f(x)^2 \rho(dx) 
    \quad
    \text{and}
    \quad
    \|f\|_n^2 = \frac{1}{n}\sum_{t=1}^{n}f(X_t)^2.
\end{align*}
Let $\|f\|_\infty = \sup_{x\in\mathcal{X}}|f(x)|$  be the supremum norm of $f$.
Consider a set of MDPs indexed by $t$. 
Each MDP is denoted by a tuple $\mathcal{M}_t=\langle\mathcal{X},\mathcal{A},R_t,P_t,\gamma\rangle$, where 
$\mathcal{X}$, a bounded closed subset of the $s$-dimensional Euclidean space, is a common state space; 
$\mathcal{A}$ is a common 
action space,
$R_t:\mathcal{X}\times\mathcal{A}\rightarrow\mathbb{R}$ is a task specific reward function uniformly bounded by $R_{\text{max}}$,
$P_t$ is a task specific transition kernel such that 
$P_t(\cdot|x,a)$ is a distribution over $\mathcal{X}$ for all $x\in\mathcal{X}$ and $a\in\mathcal{A}$,
and $\gamma\in(0,1)$ is a common discount factor.
Deterministic policies are denoted by $\pi:\mathcal{X}\rightarrow\mathcal{A}$.
For a given policy $\pi$, the MDP $\mathcal{M}_t$ is reduced 
to a Markov chain $\mathcal{M}_t^\pi = \langle\mathcal{X},R_t^\pi,P_t^\pi,\gamma\rangle$
with reward function $R_t^\pi(x)=R_t(x,\pi(x))$,
transition kernel $P_t^\pi(\cdot|x)=P_t(\cdot|x,\pi(x))$,
and stationary distribution $\rho_t^\pi$. 
The value function $V_t^\pi$ for MDP $t$ is defined as the unique fixed-point of the Bellman operator 
$\mathcal{T}_t^\pi:\mathcal{B}(\mathcal{X};V_{\text{max}}=R_{\text{max}}/(1-\gamma)) \rightarrow \mathcal{B}(\mathcal{X};V_{\text{max}})$,
 given by
\begin{equation*}
    (\mathcal{T}_t^{\pi}V)(x) = R_t^\pi(x) + \gamma\int_\mathcal{X} P_t^\pi(dy|x) V(y).
\end{equation*}
Let $\pi_t^*$ denote the optimal policy for $\mathcal{M}_t$.
The optimal value function $V_t^{\pi_t^*}$ for $\mathcal{M}_t$ is defined as the 
unique fixed-point of its optimal Bellman operator
$\mathcal{T}_t^{\pi_t^*}$
which is defined by
\begin{equation*}
    (\mathcal{T}_t^{\pi_t^*}V)(x) = \max_{a\in\mathcal{A}}\left[ R_t(x,a) + \gamma\int_\mathcal{X} P_t(dy|x,a) V(y) \right].
\end{equation*}
To approximate the value function $V$, 
we use a linear approximation architecture with 
parameters $\alpha\in \mathbb{R}^d$ and 
basis functions $\varphi_i\in\mathcal{B}(\mathcal{X};L)$ for $i=1,\cdots,d$. 
Let $\varphi(\cdot) = (\varphi_1(\cdot),\cdots,\varphi_d(\cdot))^\mathsf{T} \in\mathbb{R}^d$  be the feature vector
and  $\mathcal{F}$ the linear function space spanned by  basis functions $\varphi_i$. 
Thus, $\mathcal{F} = \{ f_{\alpha} \;|\; \alpha\in\mathbb{R}^d\; \text{and}\; f_{\alpha}(\cdot) = \varphi(\cdot)^\mathsf{T}\alpha \}$.

Consider a learning task to dynamically allocate a common resource across 
 entities $\mathcal{U}_t\subseteq\mathcal{U}$.
Each $t$  corresponds to a task, but for now 
take $t$ to be an arbitrary fixed index.
At each time step $n$, the decision maker observes  states 
$x_n = (x_{i,n})_{i\in\mathcal{U}_t}$ of the entities,
where $x_{i,n}$ is the state of entity $i$, 
and takes action $a_n = (a_{i,n})_{i\in\mathcal{U}_t}$,
where $a_{i,n}$ is the share of the resource allocated to entity $i$.
The total resource capacity is normalized to $1$ for convenience.
Therefore,  allocations satisfy  
$0\leq a_{i,n} \leq 1$ and $\sum_{i\in\mathcal{U}_t} a_{i,n} = 1$.
We consider  policy $\pi_\theta(x_n)$ parameterized by $\theta$.
Assume that we have access to the reward function $R_t$ as well as 
a simulator that  generates 
a trajectory of length $N$ given any arbitrary policy $\pi_\theta$. 
The objective of the learning task is to maximize
\begin{align*}
    J_t(\theta) = \mathbb{E}
 &   
   [   \left. \sum_{n=1}^{N} \gamma^{n-1} R_t(x_n,a_n) \;\right| 
            a_{n+1}= \pi_\theta(x_n), \; \nonumber\\
      &      x_{n+1}\sim P_t(\cdot|x_n,a_n), \;
            x_1\sim P_t(\cdot)
] 
\end{align*}
In many settings,  $N$ is small and  simulators are inaccurate;
  therefore, trajectories generated by the simulator are 
poor representations of the actual transition dynamics.
This  occurs  in  batch RL where trajectories are 
 rollouts from a dataset.
In these cases,  policies  overfit and generalize poorly.

\section{Theoretical Results}
\label{sec:theory}

We introduce first a property that we term permutation-invariance for the policy network 
that can be shown to help significantly reduce overfitting. 
\begin{definition}[Permutation Invariant Policy Network]
A policy network $\pi_\theta$ is permutation invariant 
if it satisfies $\pi_\theta(\sigma(x)) = \sigma(\pi_\theta(x))$ 
for any permutation $\sigma$.
\label{def}
\end{definition}
Permutation invariant policy networks have significant advantages over completely integrated policy networks.
While the latter are likely to fit correlations between different entities, 
this is not possible with permutation invariant policy networks as 
they are agnostic to identities of entities.
Therefore, permutation invariant policy networks are better able to
leverage experience across  time and entities, 
leading to greater efficiency in data usage.
Moreover, observe that if the transition kernels can be factored 
into independent and identical transition kernels across entities, 
then the optimal policy is indeed permutation invariant.

Our main theoretical contributions start with an extension of results from~\cite{lazaric2012finite},
where a finite-sample error bound was derived for the least squares policy iteration (LSPI) algorithm on a single task.
\cite{lazaric2012finite} provided a high-probability bound on the performance difference 
between the final learned policy and the optimal policy, of the form $c_1 + c_2/\sqrt{N}$,
where $c_1$ and $c_2$ are constants that depend on the task and the chosen feature space,
and $N$ is the number of training examples.
We extend their result by showing that, as long as tasks are $\epsilon$-close to each other 
(with respect to a similarity measure  we define later), 
the error bound of solving each task using our multi-task approach has the form $c_1 + c_2/\sqrt{NT} + c_3\epsilon$,
where $T$ is the number of tasks and $c_3$ is a task-dependent constant. 
Specifically, our theorem provides a general result and performance guarantee with respect to using data from a different but similar MDP. Definition \ref{def} provides a basis for generating many such MDPs. Finally, the benefit of doing so shall be provided by  Corollary \ref{cor:2}. 
Thus, provided $\epsilon$ is small, a given task can benefit from a much larger set of $NT$ training examples.

In addition to the assumptions  of~\cite{lazaric2012finite}, 
we extend the definition of second-order discounted-average concentrability,
 proposed in~\cite{antos2008learning},
and define the notion of first-order discounted-average concentrability.
The latter will be used in our main result, Theorem~\ref{thm:main}.
\begin{assumption}
  There exists a distribution $\mu\in\mathcal{S}(\mathcal{X})$ such that for any policy $\pi$ that is
  greedy with respect to a function in the truncated space $\tilde{\mathcal{F}}$, $\mu\leq C\rho^\pi_t$ for all $t$,
  where $C<\infty$ is a constant.
    Given the target distribution $\sigma \in\mathcal{S}(\mathcal{X})$ and
    an arbitrary sequence of policies $\{\pi_m\}_{m\geq 1}$, let
    \begin{align*}
        c_{\sigma,\mu} = \sup_{\pi_1,\ldots,\pi_m} \left\| \frac{ d(\mu P^{\pi_1}\ldots P^{\pi_m})}{d\sigma}\right\| .
    \end{align*}
    We assume that $C_{\sigma,\mu}', C_{\sigma,\mu}'' < \infty$, and define  first and second order discounted-average concentrability of future-state distributions as follows:
    \begin{align*}
        C_{\sigma,\mu}' 
        &= (1 - \gamma) \sum_{m\geq 0} \gamma^m c_{\sigma,\mu}(m),
        \\
        C_{\sigma,\mu}''
        &= (1 - \gamma)^2 \sum_{m\geq 1} m \gamma^{m-1} c_{\sigma,\mu}(m).
    \end{align*}
\end{assumption}
\begin{theorem}[Multi-Task Finite-Sample Error Bound]
    \label{thm:main}
    Let $\mathcal{M} = \langle \mathcal{X},\mathcal{A},R,P,\gamma\rangle$ be 
    an MDP with reward function $R$ and transition kernel $P$. Assume $\mathcal{A}$ finite.
    Denote its Bellman operator by
    \begin{align*}
        (\mathcal{T}^{\pi}V)(x) = R^\pi(x) + \gamma\int_\mathcal{X} P^\pi(dy|x) V(y).
    \end{align*}
    Given a policy $\pi$, define the Bellman difference operator 
    between $\mathcal{M}_t$ and $\mathcal{M}$ to be
    $\mathcal{D}_t^\pi V = \mathcal{T}_t^\pi V - \mathcal{T}^\pi V$.
    Apply the LSPI algorithm to $\mathcal{M}$,
    by generating, at each iteration $k$, 
    a path from $\mathcal{M}$ of size $N$, 
    where $N$ satisfies Lemma 4 in~\cite{lazaric2012finite}.
    Let $V_{-1}\in\tilde{\mathcal{F}}$ be an arbitrary initial value function,
    $V_0,\cdots,V_{K-1}$ ($\tilde{V}_0,\cdots,\tilde{V}_{K-1}$) 
    be the sequence of value functions (truncated value functions) 
    generated by the LSPI after $K$ iterations, 
    and $\pi_k$ be the greedy policy w.r.t. the truncated value function $\tilde{V}_{k-1}$.
    Suppose also that 
    \begin{align*}
        \|\mathcal{D}_t^\pi V^\pi\|_\mu \leq \epsilon \; \forall \;\pi, 
        \quad \text{and} \quad
        \|\mathcal{D}_t^{\pi_k} \tilde{V}_{k-1}\|_\mu \leq \epsilon \;\forall\; k.
    \end{align*}
    Then,     for  constants $c_1$, $c_2$, $c_3$, $c_4$ that are dependent on $\mathcal{M}$,
    with probability $1 - \delta$ (with respect to the random samples):
    \begin{align*}
        \| V_t^{\pi_t^*} - V_t^{\pi_{K}} \|_\sigma
        &\leq 
        c_1 \frac{1}{\sqrt{N}} + c_2 \epsilon \sqrt{C_{\sigma,\mu}'} + c_3 \sqrt{C_{\sigma,\mu}''} + c_4.
    \end{align*}
\end{theorem}
The proof is deferred to the Appendix.
Theorem~\ref{thm:main} formalizes the trade off between 
drawing fewer samples from the exact MDP $\mathcal{M}_t$,
versus drawing more samples from a different MDP $\mathcal{M}$. 
Importantly, it  shows how to benefit from solving a different MDP, $\mathcal{M}$, when:
(a) additional samples can be obtained from $\mathcal{M}$, and
(b) $\mathcal{M}$ is not too different from $\mathcal{M}_t$.
In particular, the distance measure is simply the 
distance between the Bellman operators of the MDPs, which can be bounded if the difference in
both the transition and reward functions are bounded.

    In recent work, a performance bound for multi-task learning was given in
    Theorem 2 and 3 of~\cite{deramo2020learning}.
    However, the authors used a different setup containing
    both shared and task-specific representations, 
    and their focus was on showing that the cost of learning the shared representation
    decreases with more tasks.
    They did not show how the similarity or difference across tasks 
    affects performance.
    In contrast, our setup does not contain task-specific representations,
    and our focus is on how differences across  MDPs 
    impact the benefit of having more tasks (and consequently more samples).
    We show this in Corollary~\ref{cor:main} and Corollary~\ref{cor:2}.
\begin{remark}
  While our theoretical results are based on LSTD and LSPI and assume finite action space, our approach
  is applicable to a wide range of reinforcement learning algorithms, including policy gradient methods and
  to MDPs with continuous action spaces. Deriving similar results for a larger family of models and
  algorithms remains
  an interesting, albeit challenging, future work.
\end{remark}
Permutation invariant policy networks  allow   
using data from the global set of entities $\mathcal{U}$.
Since the policy network is agnostic to the identities of the entities, 
one can learn a single policy for all tasks, where each task $t\in[T]$
is a resource allocation problem over a subset of entities $\mathcal{U}_t$.
For notational simplicity, assume that all tasks have the same number of entities, 
and all   trajectories are of equal length $N$.
Our approach can, however, be readily extended to  tasks with different numbers of entities 
and different trajectory lengths.
Permutation invariance  allows a large set of MDPs  to leverage the result of Theorem \ref{thm:main}. In the next section we shall  provide an algorithm, motivated by the following corollaries,
and a prioritized sampling strategy for this setting that drives significantly greater sample efficiency for the original task.  The sampling strategy also helps to stabilize the learning process, reducing the risk of deleterious effects of the multi-task setting, as discussed by \cite{teh} and addressed in works such as \cite{popart, attentive}.

\begin{cor} 
    \label{cor:main}
    Let $[T]$ be a set of similar tasks such that their distance from 
    the average MDP, given by
    \begin{align*}
        (\mathcal{T}^\pi V)(x)
        &=
        \frac{1}{T}\sum_{t=1}^{T} R_t^\pi(x)
        +
        \gamma\int_\mathcal{X} \frac{1}{T}\sum_{t=1}^{T} P_t^\pi(dy|x) V(y),
    \end{align*}
    is bounded by $\epsilon$ as defined in Theorem~\ref{thm:main}.
    Let $N$ be the number of samples available in each task. 
    Let $\pi_K$ be the policy obtained at the $K$\textsuperscript{th} iteration 
    when applying LSPI to the average MDP.
    Then, the suboptimality of the policy on each task
    is $O(1/\sqrt{NT}) + O(\epsilon) + c$ for some constant $c$
    (where suboptimality is defined according to Theorem~\ref{thm:main}). 
\end{cor}
Recall that each task is formed by selecting a subset $\mathcal{U}_t$ of entities from the global set $\mathcal{U}$. We thus have the following sample gain that can be attributed to the permutation invariance of the policy network.
\begin{cor} [Sample Efficiency from Permutation Invar.]
    Let $M=|\mathcal{U}|$ and $m=|\mathcal{U}_t|$. Given fixed $M$ and $m$, there are
  $T={M \choose m}\geq\left(\frac{M}{m}\right)^m$ different tasks.  Then, by Cor. \ref{cor:main},  assuming  all pairs of tasks are weakly correlated,  the potential gain in sample efficiency is exponential in $m$.
  \label{cor:2}
    \end{cor}
  Disregarding correlation
  between samples from tasks with overlapping entities
Corollary~\ref{cor:main} and Corollary~\ref{cor:2} together suggest that
  the (up to) exponential increase in the number of
  available tasks can significantly improve sample efficiency as compared to learning each task
  separately.

  \section{Exploiting Permutation Invariance through Multi-task
    Reinforcement Learning}

  Our approach to exploiting permutation invariance is via multi-task reinforcement learning, where
  each ``task'' corresponds to a particular choice of subset $\mathcal{U}_t\subset\mathcal{U}$.
  Furthermore, for each task, we enforce permutation invariance among the entities $i$ by forcing
  the neural network to apply the same sequence of operations to the state input $x_i$ of each
  instrument through parameter sharing.
  
The proposed method, shown in Algorithm~\ref{alg:main},
 learns a single policy by sampling subsequences of trajectories from the different MDPs.
At each step, we sample a task $t$ according to 
a distribution defined by  task selection policy $p$.
Then, a minibatch sample $\mathcal{B}_t$ is drawn from the replay buffer for task $t$,
and gradient descent is performed  using the sampled transitions $\mathcal{B}_t$ 
(alternatively, samples can  be generated using policy rollouts for the specific task).
Separate replay buffers  maintained for each task  
 are updated only when the corresponding task is being used.

In contrast with other active sampling approaches in multi-task learning,
our approach maintains an estimate of the difficulty of each task $t$ 
as a score, $s_t$. 
After each training step, we update the score for only the sampled task 
based on  minibatch $\mathcal{B}_t$,  avoiding  evaluation over all the tasks.
The scoring functions depend on the sampled minibatch; 
to reduce  fluctuations in scores for each task, 
exponential smoothing is applied  $s_t \leftarrow \gamma s_t + (1 - \gamma) \cdot \mathsf{scorer}(\mathcal{B}_t) $. 
We propose a stochastic prioritization method
that interpolates between pure greedy prioritization and uniform random sampling.
Our approach is similar to prioritized experience replay (PER) by~\cite{schaul2016per},
but while classical PER prioritizes samples, 
we prioritize tasks.
The probability of sampling task $t$ is
$p_t = s_t^\alpha / \sum_{t'}s_{t'}^{\alpha}$,
where the exponent $\alpha$ determines the degree of  prioritization,
with $\alpha = 0$ corresponding to the uniform case.
We correct for bias with importance-sampling (IS) weights
$w_t = 1 / (Tp_t)^\beta$,
that  compensate for  non-uniform probabilities if $\beta = 1$.
We normalize  weights by $1 / \max_t w_t$.
Tasks on which the reward variance is high can be interpreted as
having more challenging samples, hence reward variance can be used as a scoring function. 
\begin{algorithm}
    \caption{Prioritized Multi-Task Reinforcement Learning for Increasing Sample Efficiency}
    \begin{algorithmic}
        \STATE Initialize policy network $\pi_\theta$
        \STATE Initialize replay buffers $R_1,\ldots,R_T$
        \STATE Initialize time steps $n_1 \leftarrow 1, \ldots,n_T \leftarrow 1$
        \LOOP
        \STATE Select a task $t\sim p$ to train on
        \STATE Sample a random minibatch $\mathcal{B}_t$ of transitions $(x_n,a_n,r_n,x_{n+1})$ from $R_t$
        \STATE Update policy $\theta$ using $\mathcal{B}_t$ and chosen RL approach (correcting for bias using IS weights $w$)
        \STATE Update score $s_t \leftarrow \gamma s_t + (1 - \gamma)\cdot \mathsf{scorer}(\mathcal{B}_t)$
        \STATE Update ALL selection probabilities $p$ and IS weights $w$
        \FOR{$n = n_t,\ldots,\min\{n_t+n_e,N\}$}
        \STATE For task $t$, select action $a_n$ according to current policy and exploration noise
        \STATE Execute action $a_n$, and observe reward $r_n$ and new state $x_{n+1}$
        \STATE Store transition $(x_n, a_n, r_n, x_{n+1})$ in $R_t$
        \ENDFOR
        \STATE If $n < N$, update $n_t \leftarrow n+1$, otherwise, update $n_t \leftarrow 1$
        \ENDLOOP
    \end{algorithmic}
    \label{alg:main}
\end{algorithm}

%% file: experiment.tex
\section{Experiments}

\subsection{Synthetic data}

With the aim of validating the theory presented in Section \ref{sec:theory}, we define a synthetic example to explore the efficiency gain afforded by permutation
invariance. To do so, we control
of the deviation $\epsilon$ between any two tasks, thereby
empirically validating the main theoretical results.

Consider a resource allocation problem where the observed state $x_i$ for
each entity $i\in\{1\ldots m\}$ is a single scalar $x_i\in[0,1]$.
The action space is the probability simplex, where each action
$a=(a_1\ldots a_m)$ 
indicates the fraction of  resource allocated to each entity.
The reward function is 
\[
R(x,a):=\sum_i x_i a_i -\beta_i a_i\log a_i
\]
where $\beta_i$ is a weight parameter for each entity. Note that
when $\beta_i=\beta$ for all $i$, the reward function
becomes $R(x,a)=(\sum_i x_i a_i)+\beta H(a)$ where $H$ is the
Shannon entropy. This implies that maximizing the reward involves a
tradeoff between focusing resources on high $x_i$ or distributing them
uniformly across all $i$. Note that the reward function
is permutation invariant, but that when we allow a varying $\beta_i$ over the
entities, the function deviates from being perfectly permutation invariant.
We use the range $\max_i\beta_i - \min_i\beta_i$ as a stand-in for $\epsilon$.
\begin{figure}
    \includegraphics[width=0.33\textwidth]{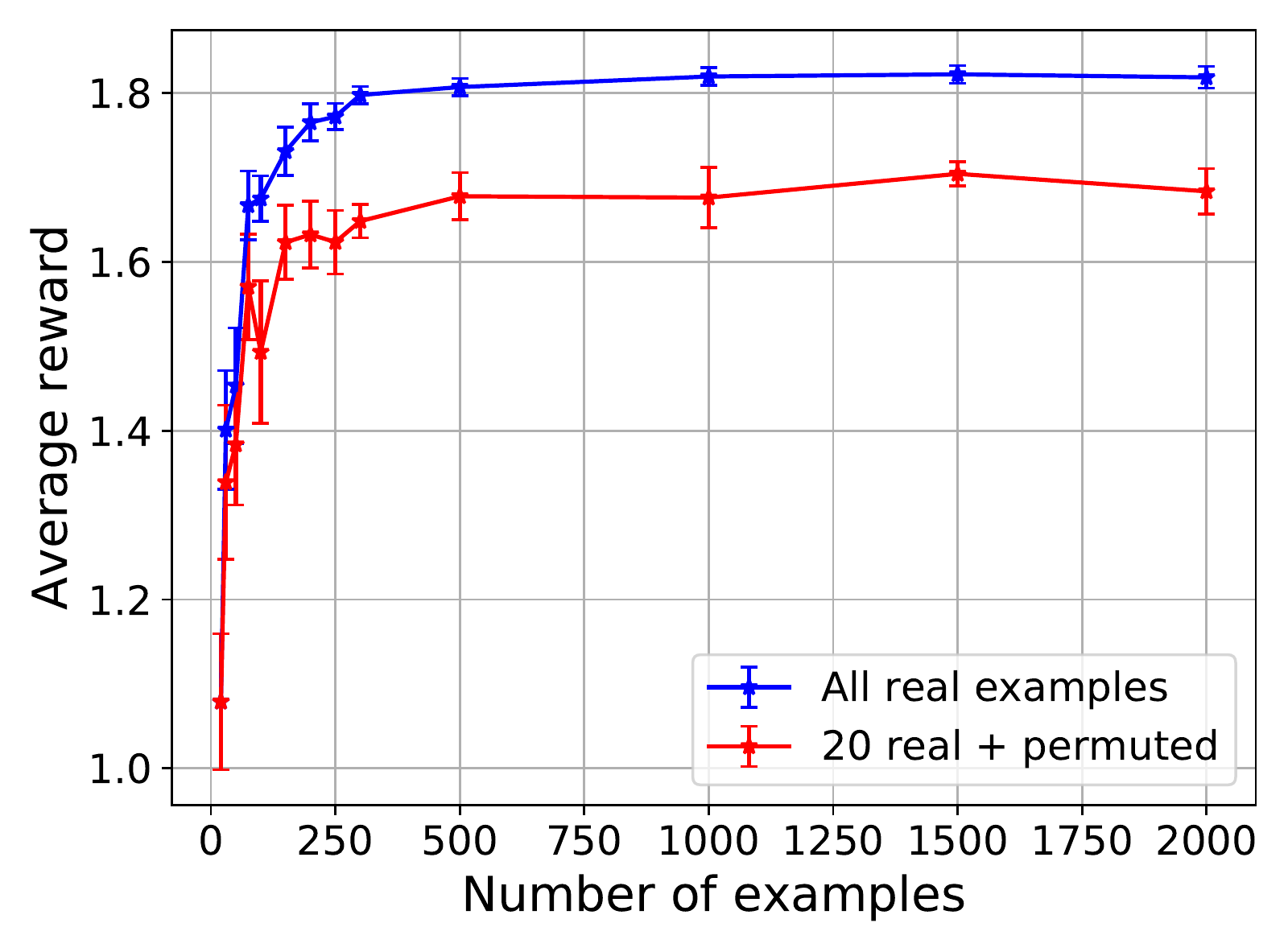} 
    \includegraphics[width=0.33\textwidth]{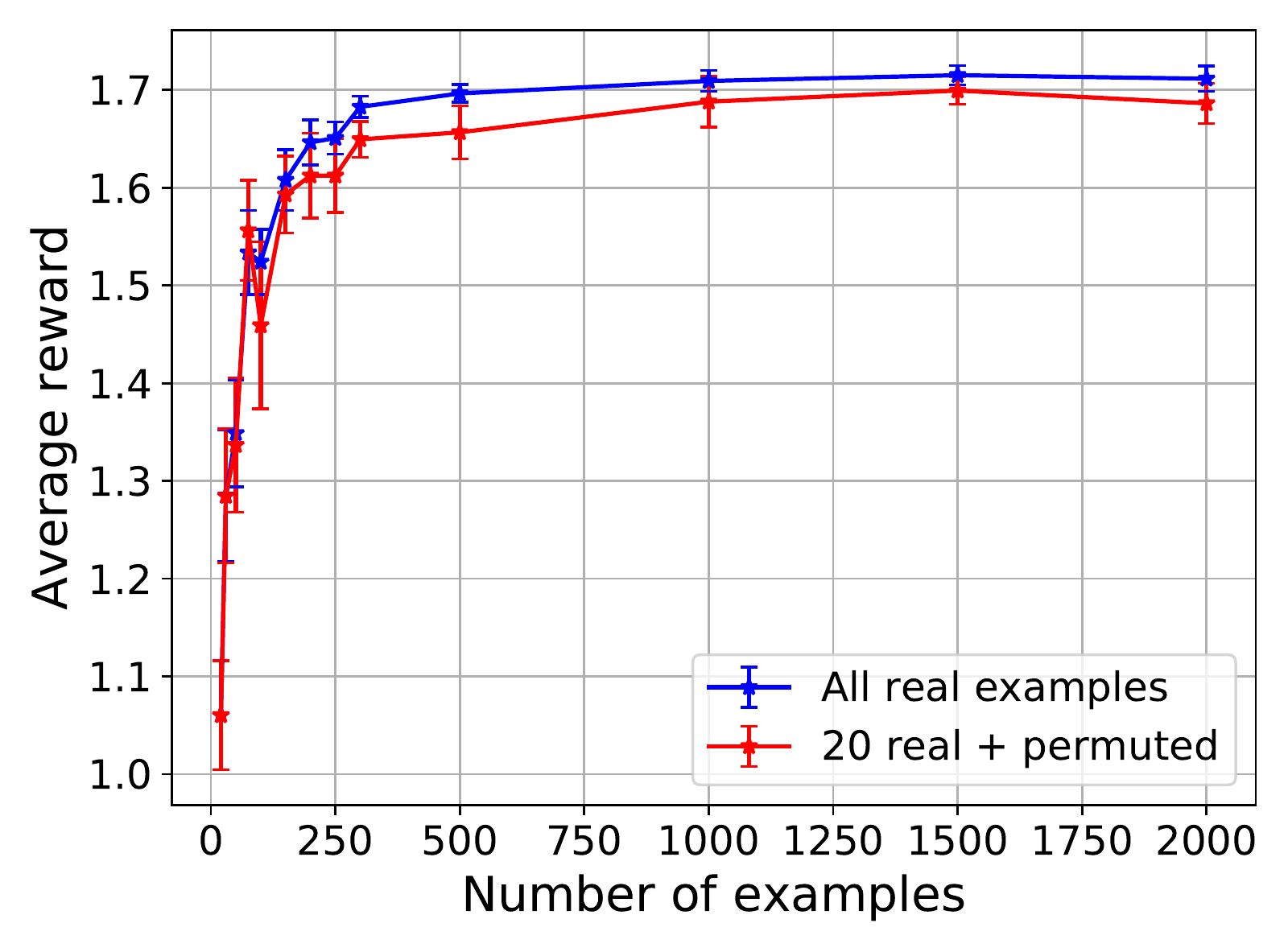}
    \includegraphics[width=0.33\textwidth]{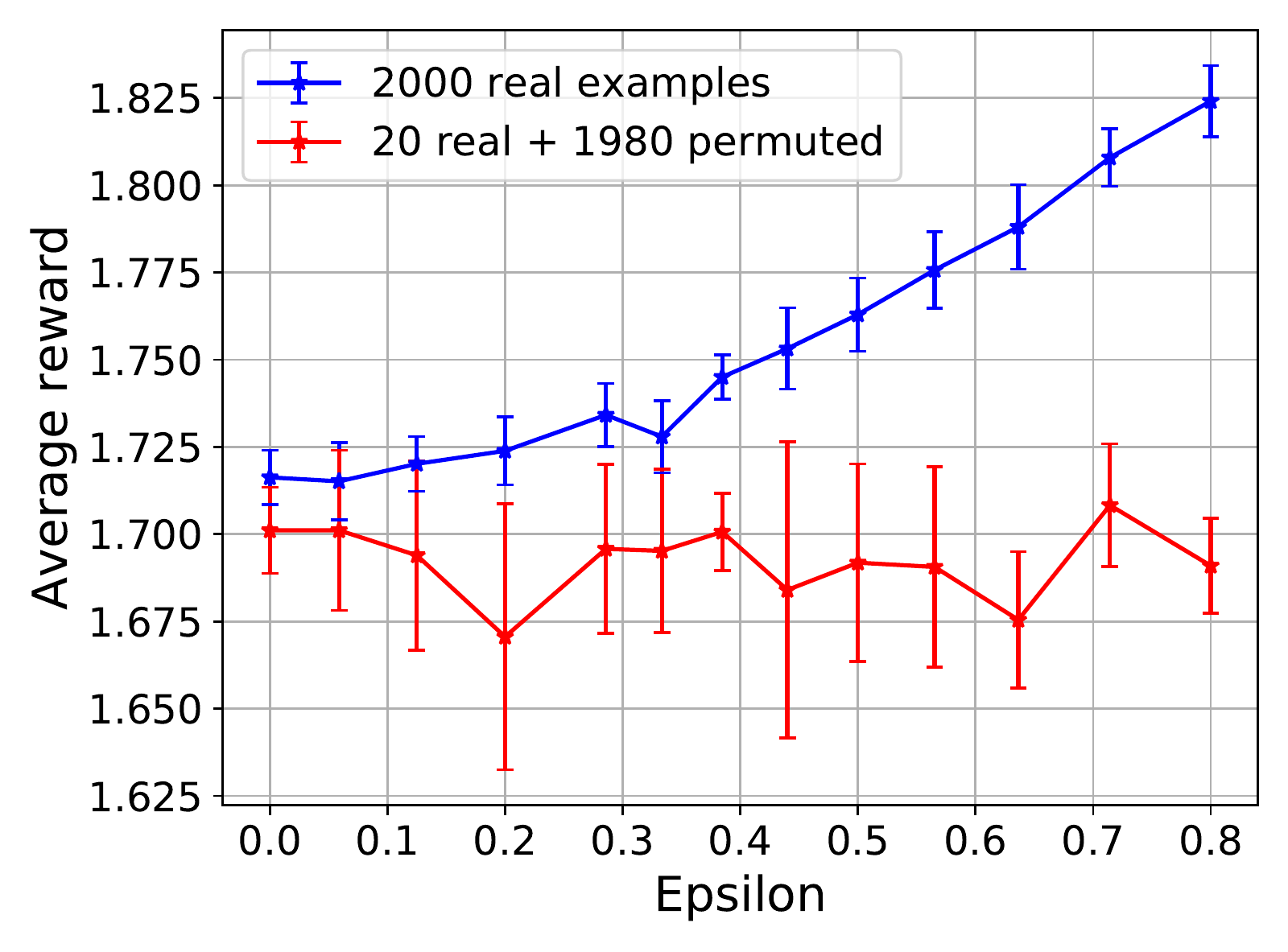}
    \caption{Performance for $\epsilon=0.8$ (left),  
    $\epsilon=0$ (middle), and at $N=2000$ with varying $\epsilon$ (right).}
    \label{fig:synthetic}
\end{figure}
Let $m=10$.
For each $\epsilon$, we run two experiments. The first  examines
 the performance of  policies trained by $LSPI$ using $N$ real
examples drawn i.i.d from the state-action space, for $N=20 \ldots 2000$.
A small Gaussian noise is added to each reward to make learning harder.
The second experiment uses only $20$ real examples, but 
augments the training set (up to $N$) through random permutation of  the real
examples.
The first two figures in Fig.~\ref{fig:synthetic} show the
results for $\epsilon=0.8$ and $\epsilon=0,$ respectively.
Performance improves with $N$, as   predicted by
the $1/\sqrt{N}$ term in our error bound. Note that
 in the experiment using only $20$ real examples, a performance gain is achieved
by using permuted examples; this corresponds precisely to the multi-task gain
 predicted by the $1/\sqrt{NT}$ term.
When $\epsilon$ is large, there is a significant gap
between the results of the two experiments, as predicted
by
the $\epsilon$-term in the error bound. The last plot
in Fig.~\ref{fig:synthetic} shows this gap at $N=2000$ when  
$\epsilon$ varies from 0 to $0.8$.

\subsection{Real-world data}

We consider two real-world resource allocation settings:
financial portfolio optimization and meta federated learning.
Financial portfolio optimization is discussed below while 
 meta federated learning is in the Appendix.
Given historical prices for a  universe of financial assets, $\mathcal{U}$,
the goal of  task $t$ is to allocate investments across
a subset of assets $\mathcal{U}_t\subseteq\mathcal{U}$. The multiple tasks $t$ thus correspond to multiple portfolios of instruments. Permutation invariance will be of use in this setting since, from a given universe of instruments (e.g. the 500 instruments in the S\&P 500), an exponential number of tasks can be generated, each with its own portfolio. Consider now one such task.

\begin{figure}
 \begin{minipage}[t]{0.49\textwidth}
        \includegraphics[width=0.49\textwidth]{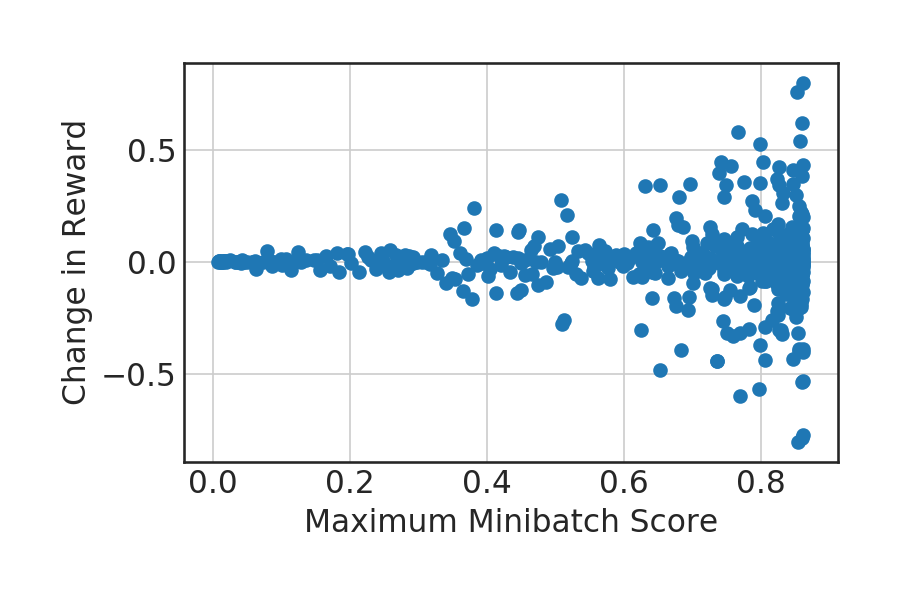}
        \includegraphics[width=0.49\textwidth]{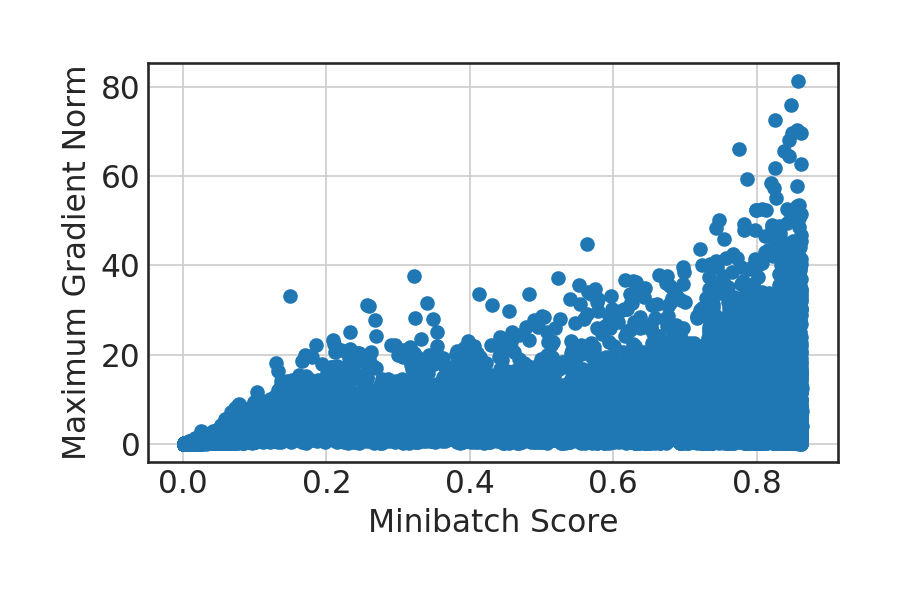}
        \caption{Scatter plots of the maximum absolute deviation from Equal CRP vs.
            the change in rewards  every $50$ steps (left)
            and the max. norm of the gradient for the minibatch (right).}
        \label{fig:active}
    \end{minipage}
    \begin{minipage}[t]{0.49\textwidth}
        \includegraphics[width=0.98\textwidth]{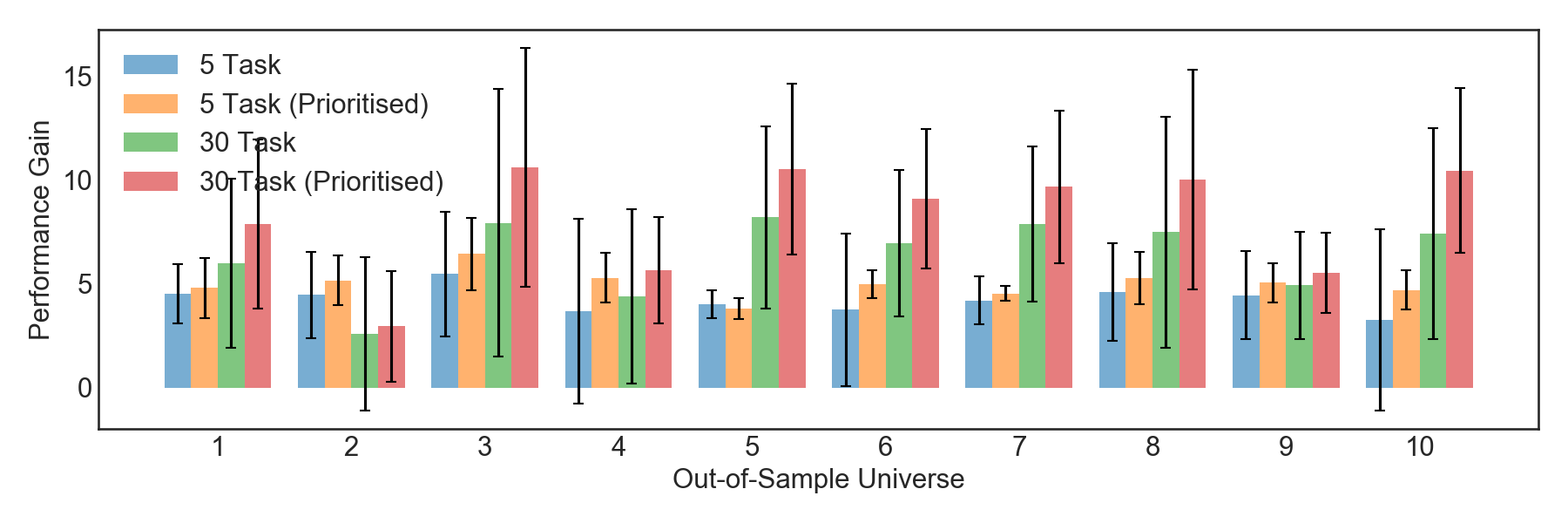}
        \caption{Mean performance gain over Equal CRP of the learned policies 
            when tested on $10$ tasks using out-of-sample instruments.
            Error bars denote the standard deviation over $10$ experiments.}
        \label{fig:transfer}
    \end{minipage}
    \hspace{10pt}
   \end{figure}

\begin{figure}
    \includegraphics[width=0.33\textwidth]{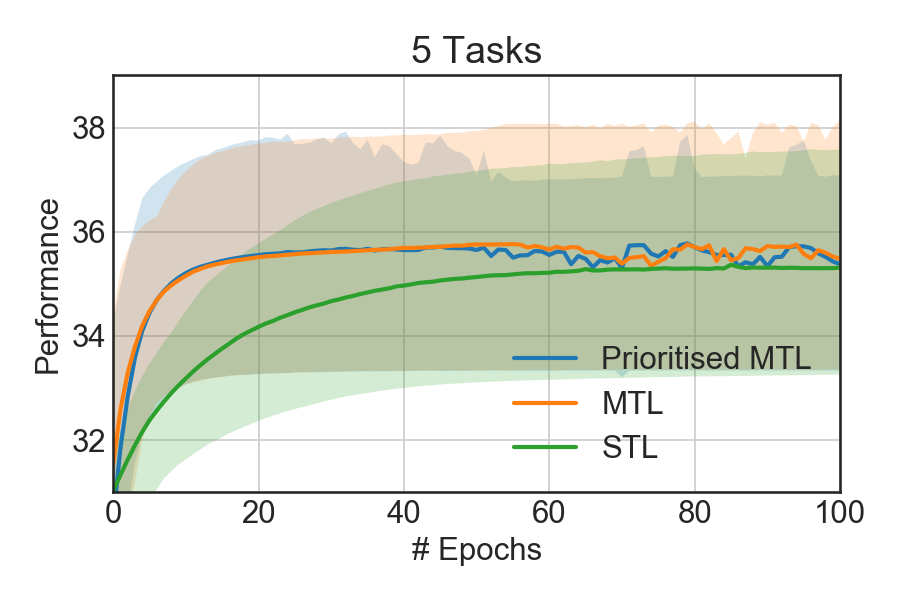} 
    \includegraphics[width=0.33\textwidth]{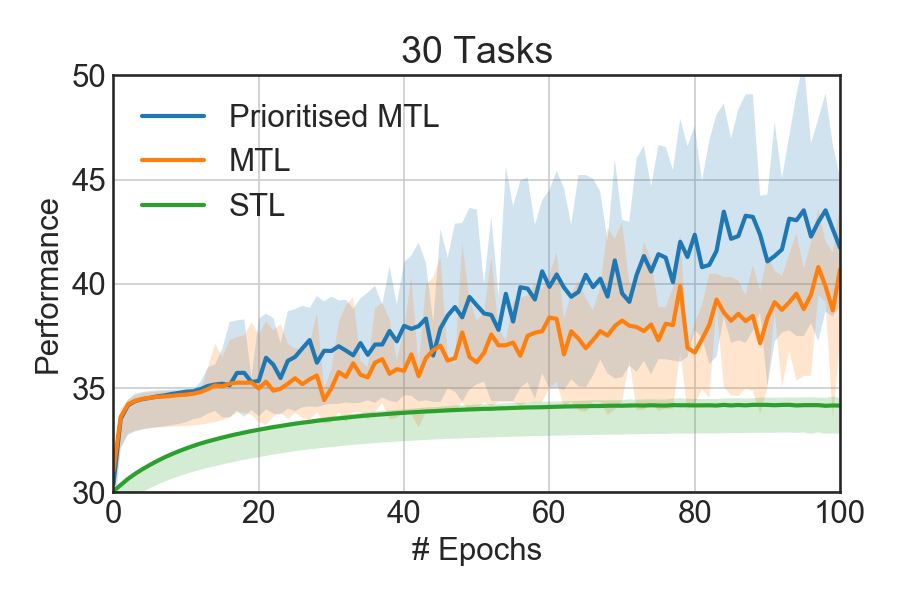}
    \includegraphics[width=0.33\textwidth]{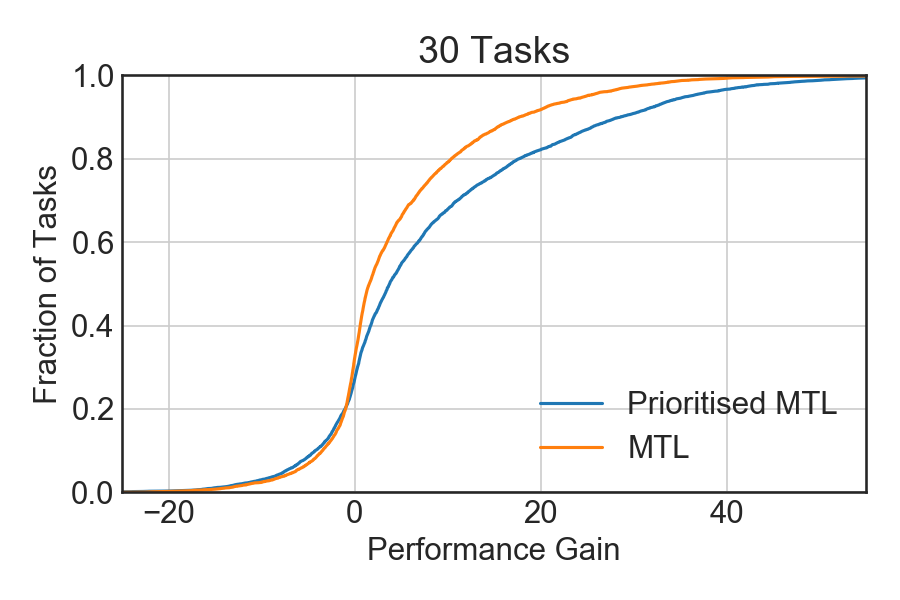}
    \caption{The two left plots show  mean annualized return in the testing period 
        over  $10$ experiments (different  instruments) each with $5$ and  $30$ tasks. 
        X-axes are scaled  to make the curves  comparable:
        each epoch has $1500$   ($5$-tasks) and $9000$ steps ($30$-tasks) and  an
        evaluation. Shaded regions denote the interquartile range.
        The rightmost figure shows, for each fraction of  tasks, the gain over Single Task Learning (STL). A curve further to the right shows higher gain over STL. From  30\% on the y-axis,  the P-MTL  gain is higher (more towards the right) than the MTL gain. As expected, when  few tasks are used, prioritizing tasks doesn't help much (y-axis from 0 to 0.2).
       }
    \label{fig:learning}
\end{figure}

At the beginning of time period $n$, the action $a_{i,n}$ 
represents the fraction of  wealth the decision maker allocates to asset $i$.
The allocations  evolve over the time period due to changes in asset prices.
Let $w_{i,n}$ denote the allocation of asset $i$ at the end of time period $n$.
We model the state of an asset using its current allocation and a 
window of its $H$ most recent prices.
In particular, let $v_{i,n}$ denote the close price of asset $i$ over time period $n$,
and let $y_{i,n} = v_{i,n} / v_{i,n-1}$ denote the ratio of close prices between adjacent time periods
\footnote{Daily high and low prices are also used in the state
but omitted here for brevity.}.
Then, the allocation in asset $i$ at the end of time period $n$ is given by
\begin{equation*}
    w_{i,n}
    =
    \frac{a_{i,n} y_{i,n} }
        {\sum_{i\in\mathcal{U}_t} a_{i,n} y_{i,n}},
\end{equation*}
and the state of asset $i$ at the beginning of time period $n$ is given by
\begin{equation*}
    x_{i,n} = (w_{i,n-1}, v_{i,n-H}/v_{i,n-1}, \ldots, v_{i,n-2}/v_{i,n-1}).
\end{equation*}
The change in  portfolio value over  period $n$ depends on
the  asset prices and
 transaction costs incurred in rebalancing the portfolio from $(w_{i,n-1})_{i\in\mathcal{U}_t}$ to $(a_{i,n})_{i\in\mathcal{U}_t}$.
The reward over  period $n$ is defined as the log rate of return:
\begin{equation*}
    R_t(x_n,a_n)
    =
    \ln \left[ \beta\left((w_{i,n-1})_{i\in\mathcal{U}_t}, (a_{i,n})_{i\in\mathcal{U}_t} \right)
    \sum_{i\in\mathcal{U}_t} a_{i,n} y_{i,n}
    \right]
\end{equation*}
where  $\beta$ can be evaluated using an iterative procedure (see~\cite{jiang2017deep}).
Defining the reward this way is appealing because 
maximizing  average total reward over  consecutive periods
is  equivalent to maximizing the total rate of return over the  periods.
To leverage this, we  approximate
$\beta ( (w_{i,n-1})_{i\in\mathcal{U}_t}, (a_{i,n})_{i\in\mathcal{U}_t}) \approx c \sum_{i\in\mathcal{U}_t} |w_{i,n-1} - a_{i,n}|$, where $c$ is a commission rate 
to obtain a closed-form expression for $R_t(x_n,a_n)$ (see~\cite{jiang2017deep}).
We optimize  using direct policy gradient 
on minibatches of consecutive samples
\begin{equation*}
    \theta \leftarrow \theta + \eta \nabla_\theta \left[\frac{1}{B}\sum_{n = n_b}^{n_b + B-1} 
    w_t R_t(x_n, \pi_\theta(x_n)) \right],
\end{equation*}
where $n_b$ is the first time index in the minibatch,
$B$  the size of a minibatch, 
and $w_t$  the IS weight for task $t$. 
As in~\cite{jiang2017deep}, we sample $n_b$ from a geometric distribution
that prioritises  recent samples and  implement replay buffers for each task.
A  benchmark trading strategy is equal constantly-rebalanced portfolio (CRP) that  rebalances to maintain equal weights.
As we noted earlier, ideally one would prefer for
the scoring function to depend only on the minibatch $\mathcal{B}_t$.
A deviation from Equal CRP can be viewed as \textit{ learning to exploit price movements}, and is thus  here we use this as the   goal of the policy.
 \textit{Prioritised MTL}  thus prioritises tasks which  deviate from Equal CRP. Note that the policy deviates from CRP  only when profitable. Let
\begin{equation*}
    \mathsf{scorer}(\mathcal{B}_t)
    =
    \max_{n\in\{n_b,\ldots,n_b+B-1\}}
    \left\| \pi_\theta(x_n) - \frac{1}{|\mathcal{U}_t|} \right\|_\infty,
\end{equation*}
be the scoring of tasks in Prioritised MTL using  mean absolute deviation 
of the minibatch  allocation from  Equal CRP.
Figure~\ref{fig:active} (left) shows a scatter plot
of the maximum score seen  every $50$ steps and 
the change in episode rewards in a single-task learning experiment, and (right) of the minibatch score and 
the maximum gradient norm  for the minibatch. 
Higher scores imply higher variance in the episode rewards
and hence more challenging and useful samples. 
The correlation between  scores and  gradient norms
shows that our approach is 
 performing gradient-based prioritisation, 
(see \cite{angelos,loshchilov2015online,alain2015variance})
but in a  computationally efficient manner. The details of the dataset and parameter settings can be found in the Appendix. Figure~\ref{fig:transfer} shows the performance of the learned policies 
 tested on $10$ tasks drawn from out-of-sample instruments. 
The policy network with weights initialized close to zero
behaves like an Equal CRP policy.
As noted, any profitable deviation from Equal CRP implies learning  useful trading strategies.
The plots show that the MTL  policies perform well on
instruments never seen during training, offering a remarkable benefit for using RL in the design of trading policies.
Fig.~\ref{fig:learning} shows the performance of prioritised multi-task learning (MTL) 
versus single-task learning (STL) (i.e. learning a policy for each task 
independently on the instruments in the task).
We also show results for MTL without prioritised sampling, i.e., with $\alpha = 0$.
We consider  $5$ tasks and  $30$ tasks.
The plots show that prioritised MTL performs significantly better 
than STL in both convergence time and final achieved performance.
The performance with $30$ tasks is significantly better 
than the performance with $5$ tasks, showing that our approach
 leverages  the  samples of the additional tasks.

\begin{figure}
    \includegraphics[width=0.33\textwidth]{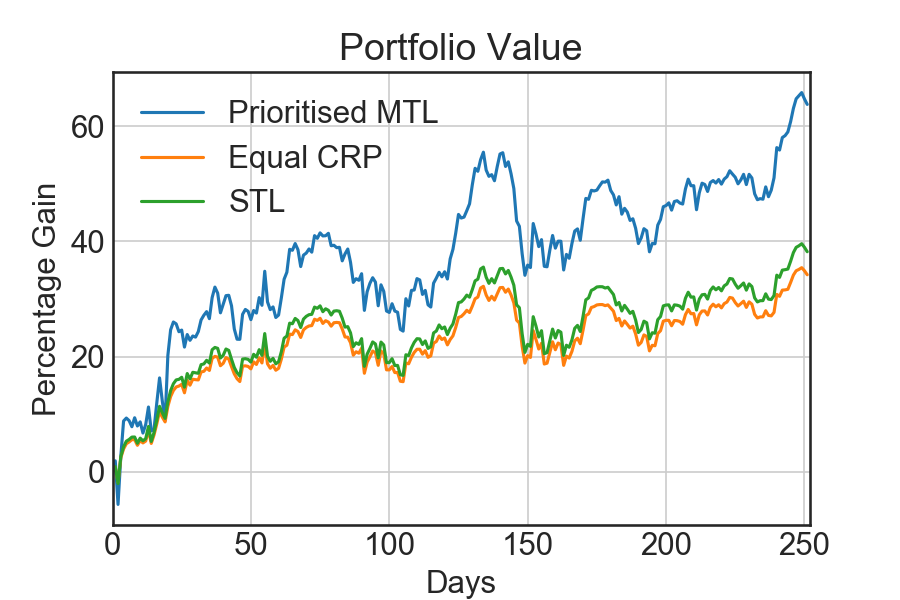} 
    \includegraphics[width=0.33\textwidth]{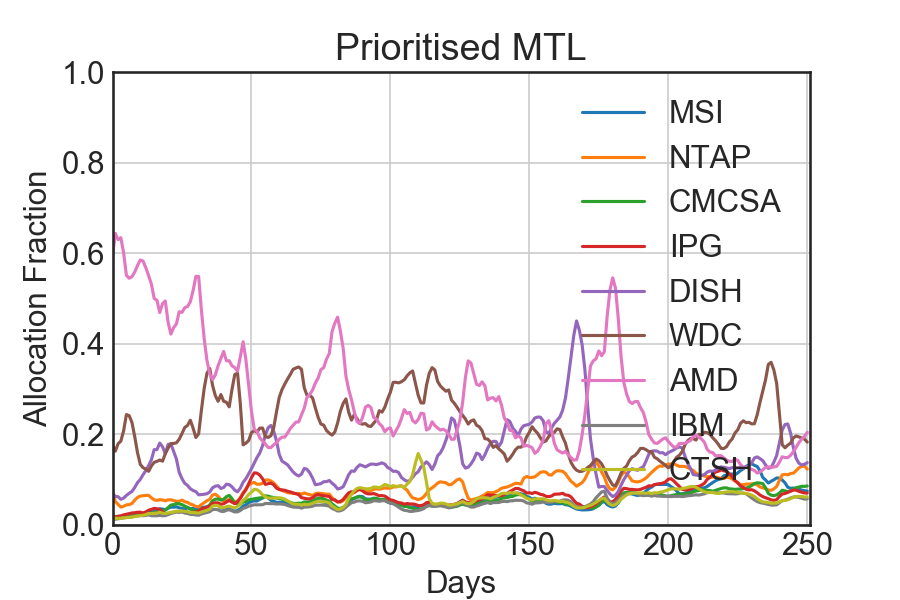}
    \includegraphics[width=0.33\textwidth]{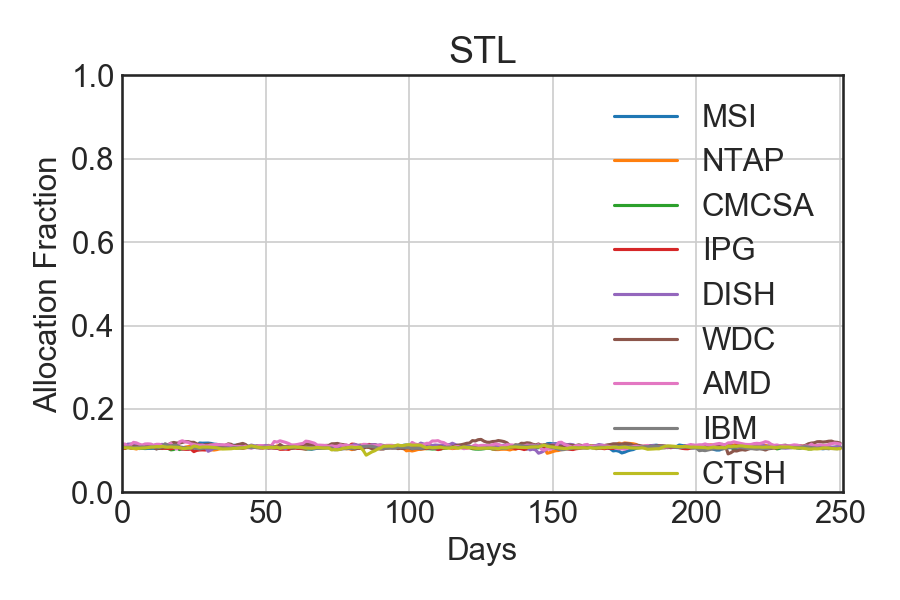}
    \caption{Comparison of a multi-task policy vs. a single-task policy 
        on the testing period for a specific task.
        The leftmost plot shows the percentage gain in portfolio value over time
        for both policies against that from the  baseline Equal CRP policy.
        The right two plots show the asset allocations.}
    \label{fig:rollout}
\end{figure}

Fig.~\ref{fig:rollout} illustrates the typical behavior of a multi-task learning (MTL)  and a single-task learning (STL) policy 
on the test period for tasks where  multi-task policy performed significantly better.
The single-task policy kept  constant equal allocations while
the multi-task policy was able to learn more complex  allocations.
In financial data,
strongly trending prices do not occur often  
and are inherently noisy.
Multi-task learning with  permutation invariance  helps with both challenges, allowing the algorithm to learn more complex patterns in a given  training period.

%% file: supp.tex

\paragraph*{Theorem 1}
    Let $\mathcal{M} = \langle \mathcal{X},\mathcal{A},R,P,\gamma\rangle$ be 
    an MDP with reward function $R$ and transition kernel $P$. 
    Denote its Bellman operator by
    \begin{align*}
        (\mathcal{T}^{\pi}V)(x) = R^\pi(x) + \gamma\int_\mathcal{X} P^\pi(dy|x) V(y).
    \end{align*}
    Given a policy $\pi$, define the Bellman difference operator 
    between $\mathcal{M}_t$ and $\mathcal{M}$ to be
    $\mathcal{D}_t^\pi V = \mathcal{T}_t^\pi V - \mathcal{T}^\pi V$.
    Apply the LSPI algorithm to $\mathcal{M}$,
    by generating, at each iteration $k$, 
    a path from $\mathcal{M}$ of size $N$, 
    where $n$ satisfies Lemma 4 in~\cite{antos2008learning}.
    Let $V_{-1}\in\tilde{\mathcal{F}}$ be an arbitrary initial value function,
    $V_0,\cdots,V_{K-1}$ ($\tilde{V}_0,\cdots,\tilde{V}_{K-1}$) 
    be the sequence of value functions (truncated value functions) 
    generated by the LSPI after $K$ iterations, 
    and $\pi_k$ be the greedy policy w.r.t. the truncated value function $\tilde{V}_{k-1}$.
    Suppose also that 
    \begin{align*}
        \|\mathcal{D}_t^\pi V^\pi\|_\mu \leq \epsilon \; \forall \;\pi, 
        \quad \text{and} \quad
        \|\mathcal{D}_t^{\pi_k} \tilde{V}_{k-1}\|_\mu \leq \epsilon \;\forall\; k.
    \end{align*}
    Then, with probability $1 - \delta$ (with respect to the random samples), we have
    \begin{align*}
        \| V_t^{\pi_t^*} - V_t^{\pi_{K}} \|_\sigma
        &\leq 
        \frac{2}{(1 - \gamma)^2}
        \left\{
            (1+\gamma) \sqrt{C C_{\sigma,\mu}''}
            \left[ 
                \frac{2}{\sqrt{1 - \gamma^2}}\left( 2\sqrt{2} E_0(\mathcal{F}) + E_2 \right)
            \right.
        \right.
        \\
        &\qquad\qquad\qquad
            \left.
                + \frac{2}{1 - \gamma}\left(\gamma V_{\text{max}}L \sqrt{\frac{d}{\nu_\mu}}(\sqrt{\frac{8\log(8dK/\delta)}{N}} + \frac{1}{N})\right) + E_1
            \right]
        \\
        &\qquad\qquad\qquad
            +
            \gamma^{\frac{K-1}{2}}R_{\text{max}}
            +
            3 \epsilon \sqrt{2 C_{\sigma,\mu}'}
        \Bigg\}.
    \end{align*}

\begin{proof}
    For convenience, we will simply remove the task subscript 
    whenever we refer to variables associated with $\mathcal{M}$.
    Define 
    \begin{align*}
        d_{t}^\pi
        &=
        \mathcal{D}_t^\pi V^\pi,
        \\
        \tilde{d}_{t,k}
        &=
        \mathcal{D}_t^{\pi_k} \tilde{V}_{k-1},
        \\
        e_k
        &= \tilde{V}_k - \mathcal{T}^{\pi_k} \tilde{V}_k,
        \\
        E_k 
        &= 
        P^{\pi_{k+1}} ( I - \gamma P^{\pi_{k+1}} )^{-1} 
        - P^{\pi^*} ( I - \gamma P^{\pi_k} )^{-1},
        \\
        F_k
        &=
        P^{\pi_{k+1}} ( I - \gamma P^{\pi_{k+1}} )^{-1}
        +
        P^{\pi^*} ( I - \gamma P^{\pi_k} )^{-1}.
    \end{align*}
    From the proof of Lemma 12 in~\cite{antos2008learning}, we get
    \begin{align*}
        V^{\pi^*} - V^{\pi_{K}}
        \leq
        \gamma \sum_{k=0}^{K-1} (\gamma P^{\pi^*})^{K-k-1} E_k e_k + (\gamma P^{\pi^*})^K ( V^{\pi^*} - V^{\pi_0} ).
    \end{align*}
    By applying the above inequality, and taking the absolute value on both sides point-wise, we get
    \begin{align*}
        &| V_t^{\pi_t^*} - V_t^{\pi_{K}} |
        \\
        &=
        | V_t^{\pi_t^*} - V^{\pi^*} | 
        + | V^{\pi^*} - V^{\pi_{K}} | 
        + | V^{\pi_{K}} - V_t^{\pi_{K}} |
        \\
        &\leq
        \gamma \sum_{k=0}^{K-1} (\gamma P^{\pi^*})^{K-k-1} F_k |e_k|
        + (\gamma P^{\pi^*})^K | V^{\pi^*} - V^{\pi_0} |
        + |V_t^{\pi_t^*} - V^{\pi^*}|
        + |V^{\pi_{K}} - V_t^{\pi_{K}}|
        \\
        &\leq
        \gamma \sum_{k=0}^{K-1} (\gamma P^{\pi^*})^{K-k-1} F_k |e_k|
        + \frac{2 R_\text{max}}{ 1 - \gamma } \gamma^K
        + |V_t^{\pi_t^*} - V^{\pi^*}|
        + |V^{\pi_{K}} - V_t^{\pi_{K}}|
    \end{align*}
    where we used the fact that $ |V^{\pi^*} - V^{\pi_0}| \leq (2 R_{\text{max}}/(1-\gamma))\mathbf{1}$.
    Next, we derive upper bounds for $|V_t^{\pi_t^*} - V^{\pi^*}|$ and $|V^{\pi_K} - V_t^{\pi_K}|$.

    \begin{itemize}
        \item[(a)] Observe that
            \begin{align*}
                V_t^{\pi_t^*} - V^{\pi^*} 
                &= 
                \mathcal{T}_t^{\pi_t^*} V_t^{\pi_t^*} - \mathcal{T}^{\pi^*} V^{\pi^*}
                \\
                &\leq 
                \mathcal{T}_t^{\pi_t^*} V_t^{\pi_t^*} - \mathcal{T}^{\pi_t^*} V^{\pi^*}
                \\
                &=
                \mathcal{T}_t^{\pi_t^*} V_t^{\pi_t^*} - \mathcal{T}^{\pi_t^*} V_t^{\pi_t^*}  
                + \mathcal{T}^{\pi_t^*} ( V_t^{\pi_t^*} - V^{\pi^*} )
                \\
                &\leq
                ( I - \gamma P_t^{\pi_t^*} )^{-1}
                d_t^{\pi_t^*}.
            \end{align*}
            The first inequality follows from the fact that 
            $\pi^*$ is optimal with respect to $V^{\pi^*}$.
            The second inequality follows from the taylor expansion of the inverse term.
            By closely following the same steps, we also get
            \begin{align*}
                V_t^{\pi_t^*} - V^{\pi^*}
                &= 
                \mathcal{T}_t^{\pi_t^*} V_t^{\pi_t^*} - \mathcal{T}^{\pi^*} V^{\pi^*}
                \\
                &\geq 
                \mathcal{T}_t^{\pi^*} V_t^{\pi_t^*} - \mathcal{T}^{\pi^*} V^{\pi^*}
                \\
                &=
                \mathcal{T}_t^{\pi^*} V^{\pi^*} - \mathcal{T}^{\pi^*} V^{\pi^*}
                + \mathcal{T}_t^{\pi^*} ( V_t^{\pi_t^*} - V^{\pi^*} )
                \\
                &\geq
                ( I - \gamma P_t^{\pi^*} )^{-1}
                d_t^{\pi^*}.
            \end{align*}
            By splitting into positive and negative components and applying the above bounds, we get
            \begin{align*}
                | V_t^{\pi_t^*} - V^{\pi^*} |
                &=
                | ( V_t^{\pi_t^*} - V^{\pi^*} )_+ - ( V_t^{\pi_t^*} - V^{\pi^*} )_- |
                \\
                &\leq
                | ( V_t^{\pi_t^*} - V^{\pi^*} )_+ | + | ( V_t^{\pi_t^*} - V^{\pi^*} )_- |
                \\
                &\leq
                | ( I - \gamma P_t^{\pi_t^*} )^{-1}
                d_t^{\pi_t^*}| + 
                | ( I - \gamma P_t^{\pi^*} )^{-1}
                d_t^{\pi^*} | 
                \\
                &\leq
                ( I - \gamma P_t^{\pi_t^*} )^{-1}
                | d_t^{\pi_t^*} | + 
                ( I - \gamma P_t^{\pi^*} )^{-1}
                | d_t^{\pi^*} | 
            \end{align*}
        \item[(b)] Observe that
            \begin{align*}
                V^{\pi_{K}} - V_t^{\pi_{K}}
                &\leq
                \mathcal{T}^{\pi_{K}}V^{\pi_{K}}
                + \mathcal{T}^{\pi_{K}}\tilde{V}_{K-1}
                - \mathcal{T}_t^{\pi_{K}}\tilde{V}_{K-1}
                - \mathcal{T}_t^{\pi_{K}}V_t^{\pi_{K}}
                \\
                &=
                \mathcal{T}^{\pi_{K}}V^{\pi_{K}}
                + \mathcal{T}^{\pi_{K}}\tilde{V}_{K-1}
                - \mathcal{T}_t^{\pi_{K}}\tilde{V}_{K-1}
                - \mathcal{T}_t^{\pi_{K}}V^{\pi_{K}}
                + \mathcal{T}_t^{\pi_{K}} ( V^{\pi_{K}} - V_t^{\pi_{K}} )
                \\
                &\leq
                ( I - \gamma P_t^{\pi_{K}} )^{-1}
                ( -d_t^{\pi_K} - \tilde{d}_{t,K} ).
            \end{align*}
            The first inequality follows from the fact that 
            $\pi_{K}$ is optimal with respect to $\tilde{V}_{K-1}$.
            The second inequality follows from the taylor expansion of the inverse term.
            By closely following the same steps, we also get
            \begin{align*}
                V^{\pi_{K}} - V_t^{\pi_{K}}
                &\geq
                \mathcal{T}^{\pi_{K}}V^{\pi_{K}}
                - \mathcal{T}^{\pi_{K}}\tilde{V}_{K-1}
                + \mathcal{T}_t^{\pi_{K}}\tilde{V}_{K-1}
                - \mathcal{T}_t^{\pi_{K}}V_t^{\pi_{K}}
                \\
                &=
                \mathcal{T}^{\pi_{K}}V^{\pi_{K}}
                - \mathcal{T}^{\pi_{K}}\tilde{V}_{K-1}
                + \mathcal{T}_t^{\pi_{K}}\tilde{V}_{K-1}
                - \mathcal{T}_t^{\pi_{K}} V^{\pi_{K}}
                + \mathcal{T}_t^{\pi_{K}} ( V^{\pi_{K}} - V_t^{\pi_{K}} )
                \\
                &\geq
                ( I - \gamma P_t^{\pi_{K}} )^{-1}
                ( -d_t^{\pi_K} + \tilde{d}_{t,K} ).
            \end{align*}
            By splitting into positive and negative components and applying the above bounds, we get
            \begin{align*}
                | V^{\pi_{K}} - V_t^{\pi_{K}} |
                &=
                | ( V^{\pi_{K}} - V_t^{\pi_{K}} )_+ -
                  ( V^{\pi_{K}} - V_t^{\pi_{K}} )_- |
                \\
                &\leq
                | ( V^{\pi_{K}} - V_t^{\pi_{K}} )_+| +
                | ( V^{\pi_{K}} - V_t^{\pi_{K}} )_-|
                \\
                &=
                | ( I - \gamma P_t^{\pi_{K}} )^{-1}
                ( -d_t^{\pi_K} - \tilde{d}_{t,K} ) |
                +
                | ( I - \gamma P_t^{\pi_{K}} )^{-1}
                ( -d_t^{\pi_K} + \tilde{d}_{t,K} ) |
                \\
                &\leq
                ( I - \gamma P_t^{\pi_{K}} )^{-1}
                | -d_t^{\pi_K} - \tilde{d}_{t,K} |
                +
                ( I - \gamma P_t^{\pi_{K}} )^{-1}
                | -d_t^{\pi_K} + \tilde{d}_{t,K} |
                \\
                &\leq
                2 ( I - \gamma P_t^{\pi_{K}} )^{-1}
                ( | d_t^{\pi_K} | + | \tilde{d}_{t,K} | ).
            \end{align*}
    \end{itemize}
    By applying the upper bounds from (a) and (b), we get
    \begin{align*}
        |V_t^{\pi_t^*} - V_t^{\pi_{K}}|
        &\leq
        \frac{2 ( 1 - \gamma^{K+2} )}{ ( 1 - \gamma )^2 }
        \left[
            \sum_{k=0}^{K-1} \alpha_{k} A_k |e_k|
            +
            \alpha ( R_{\text{max}} / \gamma )
        \right.
        \\
        &\qquad\qquad\qquad\qquad
            +
            (\beta/6) B^{\pi_t^*} \cdot 6| d_t^{\pi_t^*} |
            +
            (\beta/6) B^{\pi^*} \cdot 6| d_t^{\pi^*} |
        \\
        &\qquad\qquad\qquad\qquad
            +
            (\beta/3) B^{\pi_K} \cdot 6| d_t^{\pi_K} |
            +
            (\beta/3) B^{\pi_K} \cdot 6| \tilde{d}_{t,K} |
        \Bigg]
    \end{align*}
    where we introduced the positive coefficients
    \begin{align*}
        \alpha_k
        &=
        \frac{ ( 1 - \gamma ) }{ 1 - \gamma^{K+2} } \gamma^{K-k}, \quad \text{for}\; 0\leq k < K,
        \\
        \alpha
        &=
        \frac{ ( 1 - \gamma ) }{ 1 - \gamma^{K+2} } \gamma^{K+1},
        \\
        \beta
        &=
        \frac{ ( 1 - \gamma )}{ 2 ( 1 - \gamma^{K+2} ) },
    \end{align*}
    and the operators
    \begin{align*}
        A_k 
        &= 
        \frac{1 - \gamma}{2} (P^{\pi^*})^{K-k-1} F_k, \quad \text{for}\; 0\leq k < K,
        \\
        B^\pi
        &=
        (1 - \gamma) ( I - \gamma P_t^\pi ) ^{-1}.
    \end{align*}
    Let $\lambda_K = \left[ \frac{2 (1 - \gamma^{K+2})}{ (1 - \gamma)^2 } \right]^p$.
    Note that the coefficients $\alpha_k$, $\alpha$, and $\beta$, sum to $1$,
    and the operators are positive linear operators that satisfy
    $A_k\mathbf{1} = \mathbf{1}$ and $B^\pi\mathbf{1} = \mathbf{1}$.
    Therefore, by taking the $p$th power on both sides, 
    applying Jensen's inequality twice, 
    and then integrating both sides with respect to $\sigma(x)$, we get
    \begin{align*}
        \| V_t^{\pi_t^*} - V_t^{\pi_K} \|_{p,\sigma}^p
        &=
        \int\sigma(dx) | V_t^{\pi_t^*} - V_t^{\pi_K} |^p
        \\
        &\leq
        \lambda_K \sigma
        \left[
            \sum_{k=0}^{K-1} \alpha_{k} A_k |e_k|^p 
            + \alpha (R_{\text{max}}/\gamma)^p
        \right.
        \\
        &\qquad\qquad
            +
            (\beta/6) B^{\pi_t^*} (6| d_t^{\pi_t^*} |)^p
            +
            (\beta/6) B^{\pi^*} (6| d_t^{\pi^*} |)^p
        \\
        &\qquad\qquad
            +
            (\beta/3) B^{\pi_K} (6 | d_t^{\pi_K} |)^p
            +
            (\beta/3) B^{\pi_K} (6 | \tilde{d}_{t,K} |)^p
        \Bigg].
    \end{align*}
    From the definition of the coefficients $c_{\sigma,\mu}(m)$, we get
    \begin{align*}
        \sigma A_k 
        &\leq 
        (1 - \gamma) \sum_{m \geq 0} \gamma^m c_{\sigma,\mu}(m + K - k) \mu,
        \\
        \sigma B^{\pi} 
        &\leq
        (1 - \gamma) \sum_{m\geq 0} \gamma^m c_{\sigma,\mu}(m) \mu.
    \end{align*}
    Therefore, it follows that
    \begin{align*}
        \sigma \left[ \sum_{k=0}^{K-1} \alpha_{k} A_k |e_k|^p \right]
        &\leq 
        ( 1 - \gamma ) \sum_{k=0}^{K-1} \alpha_k \sum_{m\geq 0} \gamma^m c_{\sigma,\mu}(m + K - k) \mu | e_k |^p
        \\
        &=
        \frac{ \gamma (1 - \gamma)^2}{1 - \gamma^{K+2}} \sum_{k=0}^{K-1} \sum_{m\geq0} \gamma^{m+K-k-1} c_{\sigma,\mu}(m+K-k) \|e_k\|_{p,\mu}^p
        \\
        &\leq
        \frac{ \gamma}{1 - \gamma^{K+2}} C_{\sigma,\mu}'' e^p
    \end{align*}
    where $e = \max_{0\leq k < K}\|e_k\|_{p,\mu}^p$. The terms involving $B^\pi$ satisfy
    \begin{align*}
        \sigma \left[ B^{\pi} (6|d_t^{\pi}|)^p \right]
        &\leq
        6^p (1 - \gamma) \sum_{m\geq 0} \gamma^m c_{\sigma,\mu}(m) \mu |d_t^{\pi}|^p
        \leq
        6^p C_{\sigma,\mu}' \|d_t^{\pi}\|_{p,\mu}^p.
    \end{align*}
    Putting all these together, and choosing $p = 2$, we get
    \begin{align*}
        \| V_t^{\pi_t^*} - V_t^{\pi_K} \|_{\sigma}
        &\leq
        \lambda_K^{\frac{1}{2}}
        \left[
            \frac{\gamma}{1 - \gamma^{K+2}} C_{\sigma,\mu}'' e^2
            +
            \frac{ ( 1 - \gamma) \gamma^{K+1} }{ 1 - \gamma^{K+2} } (R_{\text{max}}/\gamma)^2
            +
            \frac{36 ( 1 - \gamma) }{2 (1 - \gamma^{K+2} ) } C_{\sigma,\mu}' \epsilon^2
        \right]^{\frac{1}{2}}
        \\
        &\leq
        \frac{2}{ (1 - \gamma)^2 }
        \left[
            \gamma C_{\sigma,\mu}'' e^2
            + (1 - \gamma)\gamma^{K+1} ( R_{\text{max}}/\gamma )^2
            + \frac{36 ( 1 - \gamma) }{2} C_{\sigma,\mu}' \epsilon^2
        \right]^{\frac{1}{2}}
        \\
        &\leq
        \frac{2}{ (1 - \gamma)^2 }
        \left[
            C_{\sigma,\mu}'' e^2
            + \gamma^{K+1} ( R_{\text{max}}/\gamma )^2
            + 18 C_{\sigma,\mu}' \epsilon^2
        \right]^{\frac{1}{2}}
        \\
        &\leq
        \frac{2}{ (1 - \gamma)^2 }
        \left[
            \sqrt{C_{\sigma,\mu}''} e
            + \gamma^{\frac{K-1}{2}} R_{\text{max}}
            + 3 \epsilon \sqrt{2 C_{\sigma,\mu}'} 
        \right].
    \end{align*}
    The desired result can then be obtained by applying the same steps as in the proof of Theorem 8 in~\cite{lazaric2012finite}.

\end{proof}

\input{meta}

%% file: meta.tex
\subsection{Financial Portfolio Optimization: Additional Details}

The dataset consists of daily prices for 
$68$ instruments in the technology and communication sectors
from 2009 to 2019.
We use   2009--2018 for training and 2019 for testing.
To validate that our approach learns common features across instruments, and thus can \textit{transfer},
we  reserve $18$  instruments not seen during training for further testing. 
 The global asset universe $\mathcal{U}$ used for training  contains $50$ instruments.
 
We construct tasks by randomly choosing a portfolio of $|\mathcal{U}_t| = 10$ instruments for each task.
We create a permutation invariant policy network by applying
the same sequence of operations to every instrument state.
That is, for each instrument, the flattened input prices are passed
through a common  RNN with 25 hidden units and tanh activation,
this output is concatenated with the latest allocation fraction of the instrument, 
and  passed through a common dense layer to produce a score.
Instrument scores are passed to a softmax function to produce allocations that sum to one.
The smoothing parameter for the scores  $\gamma = 0.2$,
 $\alpha = 0.5$ for the task prioritisation parameter and
$\beta = 1.0$ to fully compensate for the prioritized sampling bias.

\subsection{Meta Federated Learning}



Suppose we have a universe of federated learning clients $\mathcal{U}$.
The goal of task $t$ is to aggregate models in a federated learning experiment 
over a subset of clients $\mathcal{U}_t\subseteq\mathcal{U}$.
At each step $n$, the action $a_{i,n}$ represents the weight assigned to 
the supervised learning model of client $i$ in the averaging procedure.
Let $v_{i,n}$ denote the model of the client (i.e. the tensor of model parameters). 
We model the state of the client as some function of its $H$ most recent models
$x_{i,n} = f(v_{i,n-H+1},\ldots,v_{i,n})$. 
Assume that the aggregator has access to a small evaluation dataset 
that it can use to approximately assess the quality of models.
We define the reward at each step to be the accuracy of the aggregate model,
$R_t(x_n,a_n) = \mathcal{L}\left( \sum_{i\in\mathcal{U}_t} a_{i,n} v_{i,n}\right)$,
where $\mathcal{L}(v)$ is a function that provides the accuracy of a model $v$ on the evaluation dataset.
Therefore, by maximizing the total return over all time periods, 
we seek to maximize both the accuracy at the final time step 
as well as the time to convergence. We optimize the policy using Proximal Policy Optimization (PPO).

We use the MNIST digit recognition problem. 
Each client observes $600$ samples from the train dataset and trains a classifier composed of 
one $5$x$5$ convolutional layer (with $32$ channels and ReLu activation) and a softmax output layer. 
We use the same permutation invariant policy network architecture as before
with $10$ hidden units in the RNN.
We randomly select $|\mathcal{U}_t|=10$ clients for each task.
We learn using an evaluation dataset comprised of $1000$ random samples from the test dataset
and test using all $10000$ samples in the test dataset.
We fix the number of federated learning iterations to $50$.

We explore the benefit of MTL in identifying useful clients 
in scenarios with skewed data distribution.
We partition the dataset such that 
$8$ of the clients in each task observe random digits between $0$ to $5$
and the remaining $2$ clients observe random digits between $6$ to $9$. 
Therefore, for each task, $20\%$ of the clients possess $40\%$ of the unique labels.
The state of each client are the accuracies of its $H$ most recent models on the evaluation dataset.

\begin{figure}
    \includegraphics[width=0.33\textwidth]{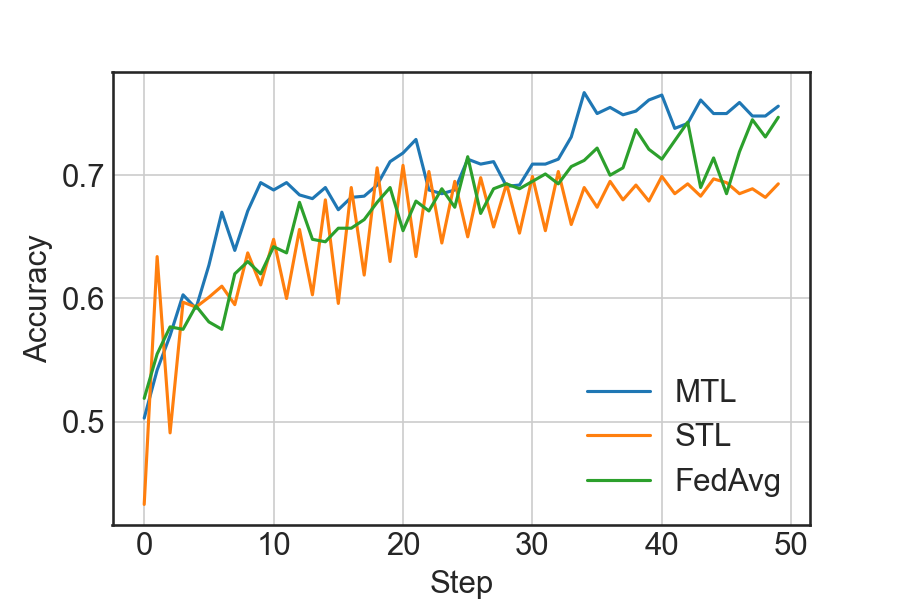} 
    \includegraphics[width=0.33\textwidth]{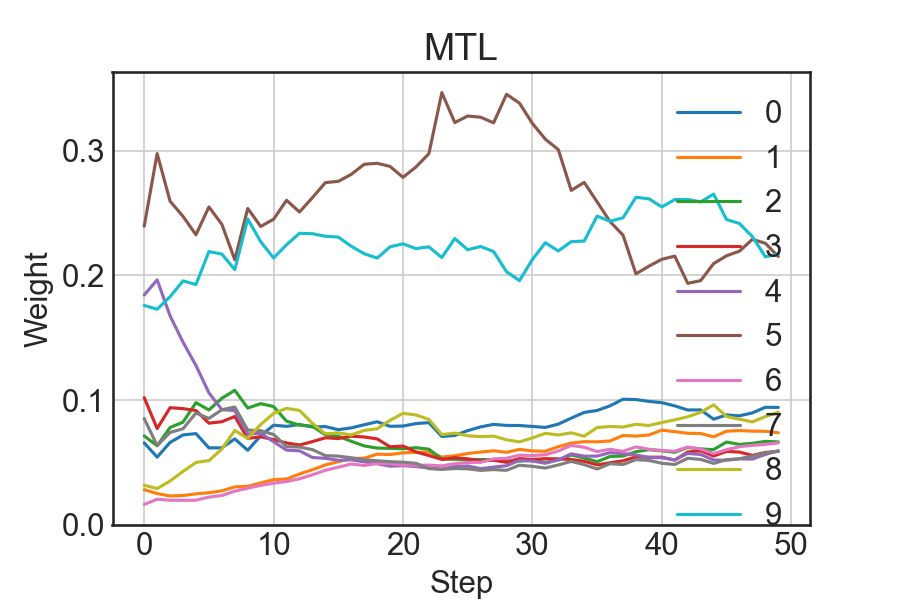}
    \includegraphics[width=0.33\textwidth]{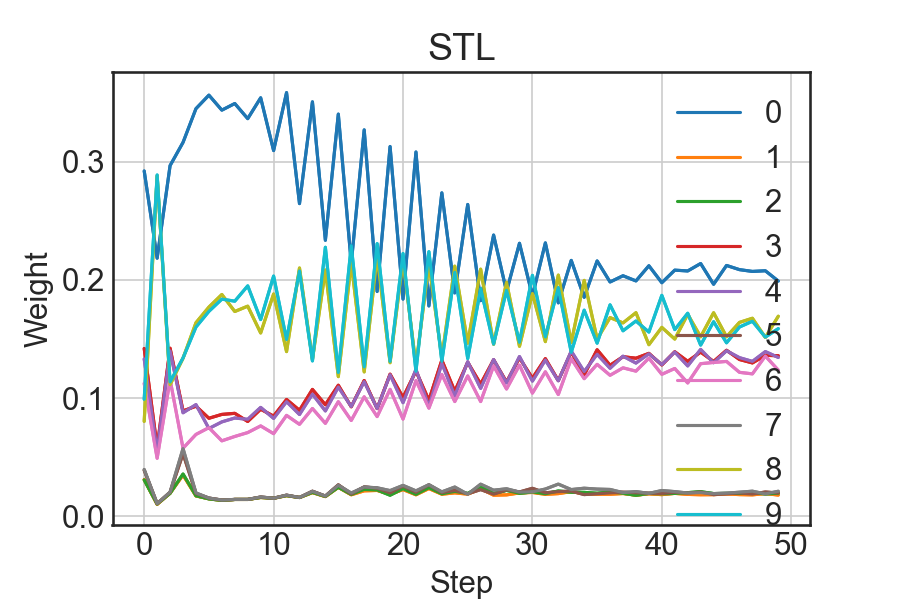}
    \caption{
        These plots compare the behavior of a multi-task policy and a single-task policy during testing. 
        FedAvg denotes the accuracy of federated learning with uniform averaging. 
        The left plot shows the accuracy of the aggregate model during federated learning.
        The right two plots show the weights produced by the policy for different clients.
        Note that clients $8$ and $9$ possess $40\%$ of the unique labels.}
    \label{fig:mnist-skew}
\end{figure}

Figure \ref{fig:mnist-skew} shows the potential benefits of multi-task learning when simulators are inaccurate.
In particular, we obtain two aggregation policies, one trained using single-task learning (STL), 
and another trained using multi-task learning (MTL), both trained using the same number of steps,
and we observe their behavior during testing.
The plots show that multi-task learning is able to learn non-uniform averaging policies 
that improve the convergence and performance of federated learning runs.
More importantly, it can perform better than single-task learning
even with the same number of samples.
This may be attributed to the wider variety of client configurations (and consequently experiences) 
in the multi-task approach.


